\documentclass[10pt,twocolumn,letterpaper]{article} 

\usepackage{avss}
\usepackage{times}
\usepackage{epsfig}
\usepackage{graphicx}
\usepackage{amsmath}
\usepackage{amssymb}
\usepackage{subcaption}
\usepackage{url}
\usepackage{fancyhdr}
\avssfinalcopy 

\def\avssPaperID{128} 

\ifavssfinal\fi

\fancypagestyle{FirstPage}{
\fancyhf{}
\lfoot{978-1-5386-9294-3/18/\$31.00~\copyright~2019 IEEE}
}

\begin{document}

\title{Automated Real-time Anomaly Detection in Human Trajectories using Sequence to Sequence Networks}
\author{Giorgos Bouritsas$^{1,2}$ \qquad Stelios Daveas$^{1}$ \qquad Antonios Danelakis$^{1}$ \qquad 
\\Constantinos Rizogiannis$^{1}$ \qquad Stelios C. A. Thomopoulos$^{1}$\\
National Center for Scientific Research ``Demokritos'', Athens, Greece$^{1}$ \\ \quad Imperial College London, UK$^{2}$\\
{\tt\small  \{gbouritsas, sdaveas, a.danelakis, crizogiannis, scat\}@iit.demokritos.gr}
}
\maketitle
\thispagestyle{FirstPage}
\begin{abstract}
Detection of anomalous trajectories is an important problem with potential applications to various domains, such as video surveillance, risk assessment, vessel monitoring and high-energy physics. Modeling the distribution of trajectories with statistical approaches has been a challenging task due to the fact that such time series are usually non stationary and highly dimensional. However, modern machine learning techniques provide robust approaches for data-driven modeling and critical information extraction. In this paper, we propose a Sequence to Sequence architecture for real-time detection of anomalies in human trajectories, in the context of risk-based security. Our detection scheme is tested on a synthetic dataset of diverse and realistic trajectories generated by the ISL iCrowd simulator \cite{Scatc,iCrowd}. The experimental results indicate that our scheme accurately detects motions that deviate from normal behaviors and is promising for future real-world applications.
\end{abstract}

\section{Introduction}\label{Sec: Intro}
Advances in video capturing devices used for video surveillance have facilitated the collection of trajectory data for moving objects, such as humans. That type of data could be used in order to extract knowledge for suspicious activity detection and real-time decision making in various fields, such as surveillance, transportation management, and border security.

The problem of detecting abnormal human behavior using trajectory data is an open research branch of computer vision. Timely detection of such events is critical for adapting mitigation measures, but it requires careful analysis of the trajectories. This is a great challenge to human operators, monitoring video surveillance footage due to information overload, which makes the whole procedure prone to errors. Thus, there is a significant need for automating the detection procedure.

In this paper, we propose a deep learning architecture for real-time detection of anomalies in pedestrian trajectories. We approach the problem using a Sequence to Sequence network (LSTM autoencoder), drawing inspiration from the impressive results that have been achieved by this architecture in the area of Machine Translation \cite{Matr1,Matr2}. It comprises of two stacked networks; the Encoder and the Decoder. The Encoder is responsible for embedding each sample in a semantically meaningful latent space, while the Decoder's goal is to reconstruct the sample leveraging the information encoded in the latent vector. The training stage is performed off-line, while the detection stage is an on-line real-time process. During the detection stage we measure the mean squared reconstruction error of each trajectory and treat it as a suspiciousness score. Anomalies are detected whenever the score exceeds a pre-defined threshold. The technique is fully automated, significantly reducing the workload of personnel monitoring video surveillance footage.  For the evaluation of the scheme, experiments on a synthetic dataset have been conducted. 
The experimental results indicate that our scheme accurately detects abnormal motion patterns and is promising for applications under real-world conditions. 


\subsection{Outline}
The rest of the paper is organized as follows: Related work is discussed in Section \ref{Sec: RelWork}. In Section \ref{Sec: DeepArch} we discuss the architecture used for detecting abnormal trajectories of pedestrians. Section \ref{Sec: Results} illustrates the evaluation of the proposed architecture as derived after appropriate experimentation. Finally,  in section \ref{Sec: FutureChallenges} and Section \ref{Sec: Conclusions} we point out future research directions and conclusions, respectively.

\section{Related Work}\label{Sec: RelWork}
Most of the state-of-the-art trajectory anomaly detection algorithms do not explicitly model the anomalous class and attempt to infer the anomaly scores based on features of normal data only.

 Initial attempts to tackle the problem resorted on handcrafting features and distance measures between trajectories in order to cluster the data points. In \cite{Li_etal_Temporal_outlier_vehicle_traffic_ICDE_2009} temporal outliers are detected by measuring historical similarity trends between data points, while in \cite{Ge_et_al_TOP-EYE_CIKM2010} the authors compute a temporally evolving score along each trajectory through a decay function. In \cite{Bu_etal_anomaly_monitoring_trajectory_streams_KDD2009} local clusters of trajectory streams are created and anomalies are monitored via efficient pruning strategies. Similarly, in \cite{Zhang_etal_iBAT_UbiComp2011,Chen_etal_iBOAT_TITN2013} a distance metric is used in order to estimate the support of various trajectory clusters. Clusters with small support are identified as anomalous. Clustering approaches based on similarity measurements are also deployed in \cite{Fu_SIMILARITY_VEHICLE_TRAJECTORY_CLUSTERING_ANOMALY_ICIP2005} and \cite{Wang_et_al_Learning_Semantic_Scene_Models_ECCV2006}. Finally, in \cite{Lee_trajectory_outlier_partition_detect_ICDE2008} the authors use a hybrid algorithm combining a distance-based and a density-based approach. 
 
 Moreover, more sophisticated methods based on probabilistic modeling and learning of normal trajectories have been used. In \cite{Suzuki_etal_motion_patterns_anomaly_human_trajectory_ICSMC2007pdf} a Hidden Markov model is deployed followed by k-means clustering. Moreover,
 Laxhammar and Falkman, in a series of works \cite{Laxhammar_Falkman_Conformal_Distribution-Independent_Anomaly_Streaming_StreamKDD2010, Laxhammar_Falkman_Conformal_Anomaly_Haussdorf_ICIF2011, Laxhammar1}, provide an anomaly detector based on conformal prediction theory that can act on incomplete trajectories in an online fashion.
 
A least popular family of techniques is based on supervised learning, either by modeling distinct classes of normal data or by casting the problem as binary classification. In \cite{Li_et_al_ROAM_SDM_2007} a rule-based classifier is implemented that applies different rules at multiple levels of granularity in order to classify each data point as normal or abnormal. In \cite{Johansson_Falkman_vessel_anomalies_Bayesian_network_ISSNIP_2007} the authors model the underlying distribution by a Bayesian network. In \cite{Liao_etal_anomaly_GPS_visual_analytics_VAST2010} and \cite{Silito_FIsher_Semi-supervised_Anomalous_Trajectory_Detection_BMCV2008} active learning is used in order for the classifier to allow for human input during training. In this way, the algorithm is is assisted to converge to a more accurate estimator of the normal data distribution. 

Most of the aforementioned methodologies are noise sensitive, need fine tuning of multiple parameters, and some of them model the trajectories as curves, thus ignoring their temporal dimension. Furthermore, statistical models are widely used (Conditional Random Fields at \cite{Liao_etal_anomaly_GPS_visual_analytics_VAST2010}, Gaussian Mixtures and Kernel Density Estimation in \cite{Laxhammar_etal_anomaly_sea_traffic_comparison_GMM_KDE_ICIF_2009}, Hidden Markov Models in \cite{Suzuki_etal_motion_patterns_anomaly_human_trajectory_ICSMC2007pdf}), instead of deep learning models. The papers that first attempted to follow the deep learning line of research for anomaly detection are \cite{newwork1} and \cite{newwork2}. However, these methods still rely on designing the input features or on a nearest neighbor search in the training set, while ours is fully automated and allows for independent classification of each data point at test time regardless of the training set.
 
\section{Sequence to Sequence Anomaly Detector}\label{Sec: DeepArch}
In our approach we assume that the trajectories are readily available, namely we can track each object's exact position for a specific time interval (similar to \cite{FINAL}). In the general case, trajectories can efficiently encode the information regarding motion in the video. In addition, potential dangerous events in crowded scenes are usually accompanied by abnormal motion (e.g. trespassing restricted areas, abrupt accelerations, loitering). The latter statements provide the motivation upon which the proposed detection scheme is based. The reduction of the workload of personnel monitoring video surveillance footage, the simultaneous processing of the behavior of multiple agents and the fact that the number of people that need to be thoroughly checked is efficiently narrowed down are some of the key benefits of the proposed methodology. 

Within our approach, trajectories are represented by a Sequence to Sequence architecture, which is realized as an LSTM autoencoder. The Encoder network embeds each sample (of arbitrary length) in a latent space in a semantically meaningful way, \ie similar trajectories will be close in the latent space, thus implicitly performing trajectory clustering. The Decoder learns to generate trajectories by using input sampled from the latent space. They system is trained off-line by minimising the distance between the input trajectory and the reconstruction provided by the Decoder. Following training, the pre-trained network can be used for on-line detection of anomalies in temporal segments of each trajectory (\ref{preprocessing}) . The detection stage contains the following steps: 1) calculation of the representation of each sub-trajectory (using the pre-trained network), 2) calculation of a suspiciousness score and 3) generation of an appropriate alert should the score exceed a pre-defined threshold. Our approach does not require explicitly describing the nature of potential anomalies (in many cases anomalies are not even strictly defined), thus rendering it appropriate for classifying arbitrary abnormal patterns. However, annotated abnormal examples can be used to refine the inference step of the algorithm, either by estimating the reconstruction error threshold or by more sophisticated techniques, such as training the autoencoder with a contrastive loss using negative samples. 

\subsection{Problem Statement}
We aim to spatio-temporally detect individuals or groups of people whose movement indicates suspicious behavior, based on the analysis of their corresponding trajectories. Throughout this paper, anomalies are defined as movements deviating from the normal motion patterns. 

Each footage contains $N(t)$ entities (e.g. humans) that move in the space captured by the visual field of the camera. Each entity $i$ follows a trajectory ${\mathcal{L}_i = \{(x_i(t), y_i(t), z_i(t))\}_{t=1}^{T_i}}$, where ${\mathbf{x}_i(t) = (x_i(t), y_i(t), z_i(t))}$, the 3D cartesian coordinates corresponding to the position of the entity $i$, as defined on a global coordinate system. The goal is to detect motion patterns (\ie trajectories) that deviate from normal behavior. The stated problem can be decomposed into two main challenges. The first one is the analysis and modeling of the structure of the normal motion patterns. The second is the detection of trajectories that do not comply with the above structure.

\subsection{Analysis and Modeling of Normal Patterns}
\subsubsection{Preprocessing}\label{preprocessing}
\begin{figure}[t]
    \begin{subfigure}{\linewidth}
    \includegraphics[width=\textwidth]
    {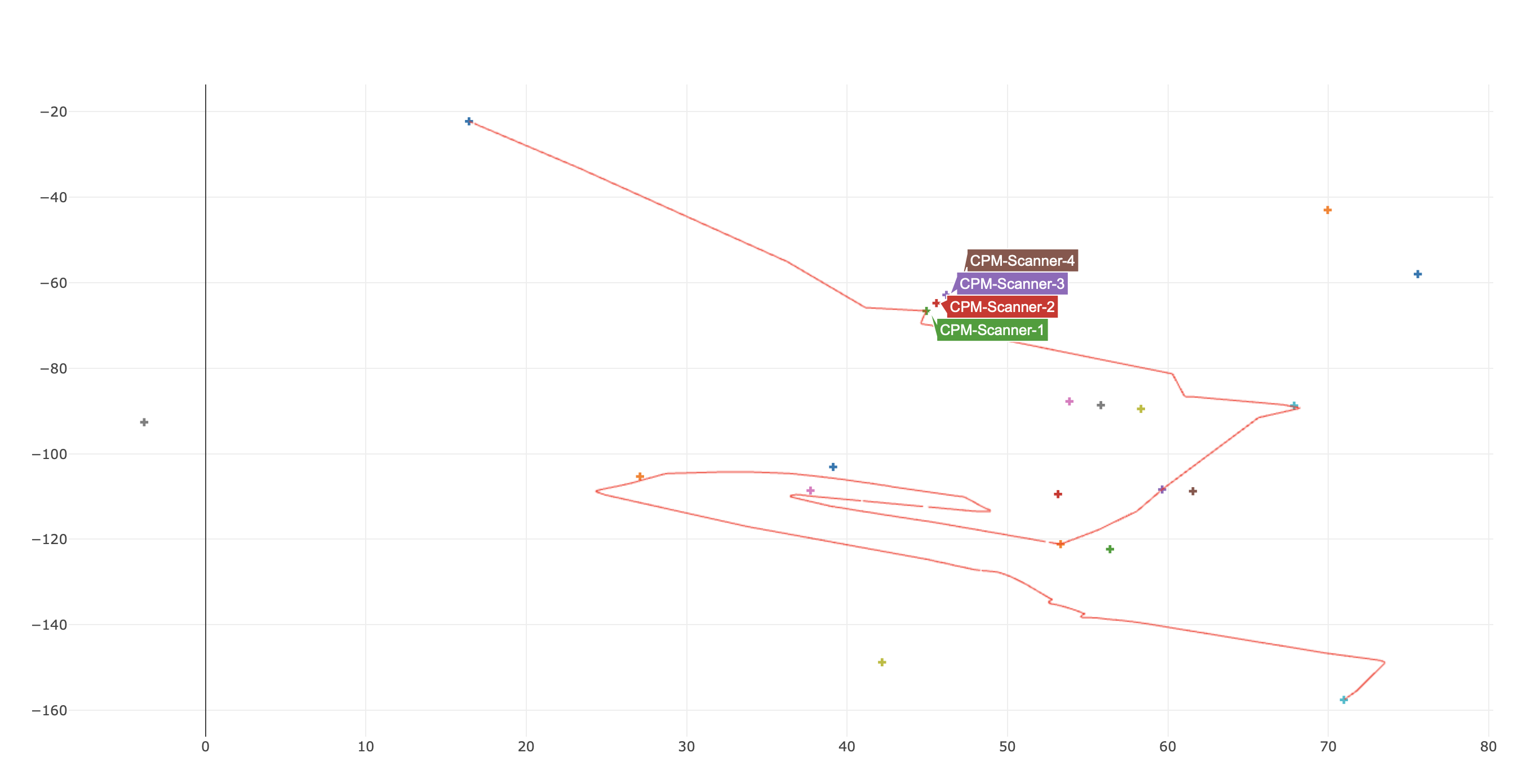}
    \label{fig:entire}
    \end{subfigure}
    \begin{subfigure}{\linewidth}
    \includegraphics[width=\textwidth]
    {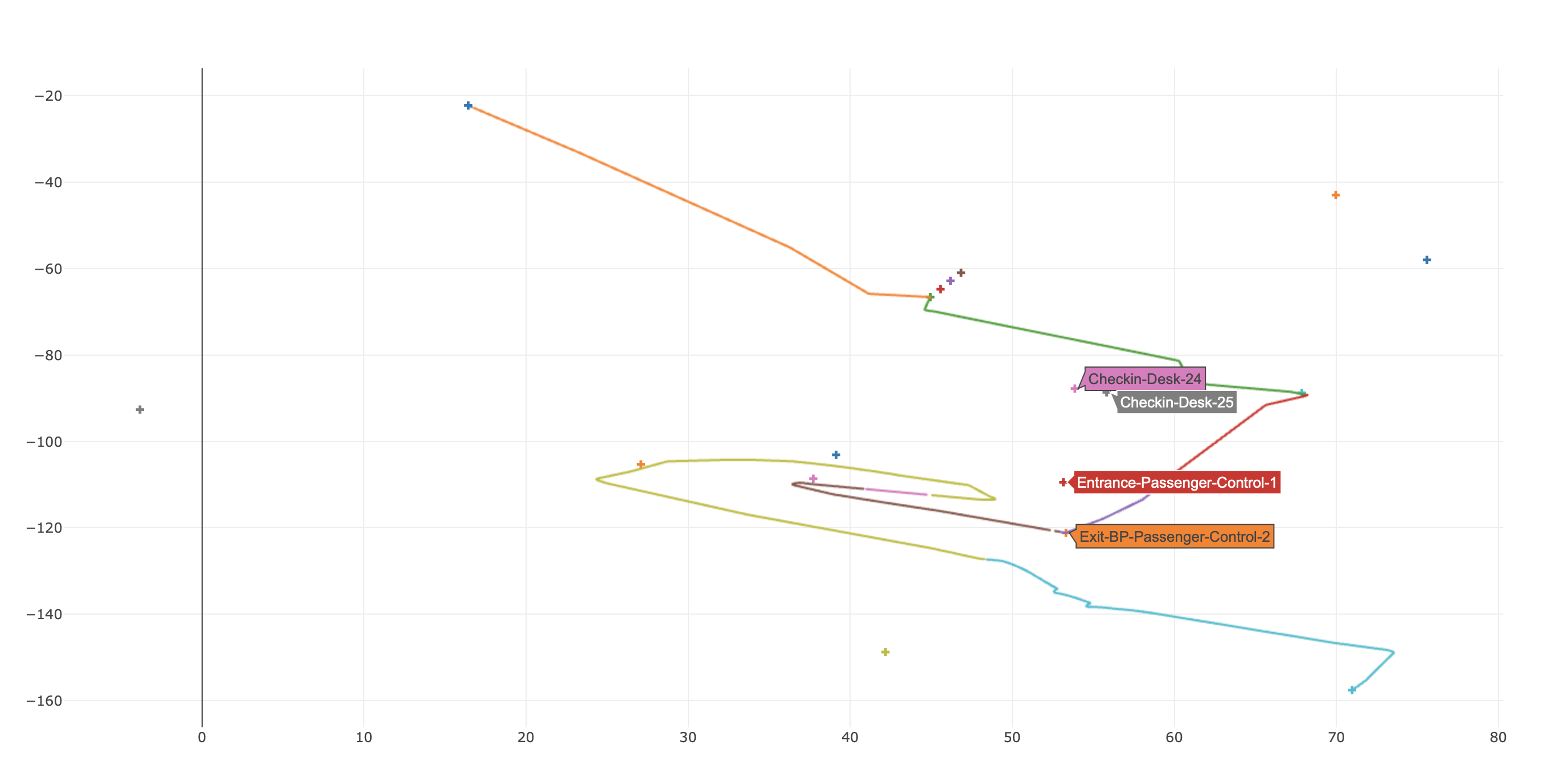}
    \label{fig:split}
    \end{subfigure}\vfill
    \caption{2D Floor plan of the airport. Upper: Entire trajectory of a passenger. Lower: Splitted to sub-trajectories based on checkpoint positions. Checkpoints are annotated with crosses. Please see color version for better visibility.}
    \label{fig:traj_split}
\end{figure}

A trajectory $\mathcal{L}_i$ may consist of various motion patterns. To allow for more accurate localization of each anomaly, we can segment each $\mathcal{L}_i$ in sub-trajectories $\mathcal{L}_i^j$. In this work, this is implemented using specific checkpoints that are provided in the dataset. Each time an entity approaches a checkpoint, its trajectory is splitted (see Figure \ref{fig:traj_split}). Initially from each sub-trajectory $\mathcal{L}_i^j$ we extract a time sequence of vectors $\mathbf{x}_i^j(t)$ with duration $T_i^j$, containing each entity's position. Note here that the network receives no information regarding the sub trajectory's characteristics (e.g checkpoint, order within the original trajectory etc.), but all the sub-trajectories are treated equally.

\subsubsection{Trajectory Modeling and Network Architecture}
Each sequence $\mathcal{L}_i^j = \{\mathbf{x}_i^j(t)\}_{t=1}^{T_i^j}$ is embedded in an $D$-dimensional space. To this end, a vector ${\mathbf{c}_i^j = f (\{\mathbf{x}_i^j(t)\}_{t=1}^{T_i^j})}$ that encodes all the spatio-temporal information that the trajectory carries is calculated. For the implementation a sequence autoencoder \cite{Implement1,Implement2} is deployed, where the building block is the Long-Short Term Memory cell \cite{hochreiter1997long}. Note here, that the trajectories might be of large lengths, depending on the frame rate of the camera, thus a model such as a LSTM is necessary in order to capture long term dependencies.

The LSTM Encoder is a Recurrent Neural Network (RNN) encoding the sequence structure in a hidden state vector $\mathbf{h}(\mathbf{x}_i^j(t))$ that is updated at each time step (fixed size, not-time dependent, much lower dimensionality). The embedding of the trajectory \textbf{(context vector)} results from the last hidden state of the Encoder, i.e $\mathbf{c}_i^j = \mathbf{h}(\mathbf{x}_i^j(T_i^j)$. The LSTM Decoder is a RNN generating a sequence, by initializing its hidden state with the context vector.

Regarding the architecture details, the Encoder is a two-layer LSTM of 64 and 32 output dimensions and the Decoder of 32 and 64 latent dimensions. A fully-connected layer maps the last LSTM output (64-dims) to a vector with dimensions equal to those of the input. We used only the position information of the entities (3-dims: x,y,z coordinates), contrary to previous methods that used handcrafted features in the input. Deeper architectures did not show any important improvement in the results, thus we used the aforementioned for the sake of simplicity and speed.

\subsubsection{Training Stage}
During the training stage, the loss function that we minimize is the mean squared $L_2$ distance shown in Equation (\ref{Eq: LossFunction}). It should be highlighted that only normal motion patterns are used as training samples.
\begin{equation}
\label{Eq: LossFunction}
    L = \frac{1}{2} \sum_{i,j}\sum_{t=1}^{T_i^j} ||\mathbf{x}_i^j(t) - D(E(\mathbf{x}_i^j(t)))||_2^2
\end{equation}

\subsubsection{Implementation Details}
We group samples into batches as follows. Steps 2 to 4 are repeated every training epoch.: 

\noindent\textbf{1) Bucketing process:} Initially, sub-trajectories with similar lengths are grouped. This is done by determining the minimum $m$ and maximum $M$ trajectory length in the dataset and then dividing the interval $[m,M]$ to equally spaced intervals (``buckets''): $\bigcup_{i}{[m_1^i,m_2^i)}$, with ${m_1^1=m}$, ${m_2^i = max(m_1^i+L,M)}$,  ${m_1^{i+1} = m_2^i}$, where $L$ is the parameter defining the size of the intervals. If a bucket contains fewer examples than a user-defined parameter $N_{min}$, then it gets concatenated to the previous bucket\footnote{We set $L$ = 100 and $N_{min}$ = 2*batch size. These parameters mainly affect the computation time of each epoch, rather than the training itself.}. 

\noindent\textbf{2) Batch generation process:} First, the examples in each bucket are shuffled. Then, the buckets are concatenated and then we traverse the dataset sequentially and group examples to batches of the same size.

\noindent \textbf{3) Padding process:} Each time sequence is padded with a masking value (a value that will be ignored during training), so that all the sequences in the batch have exactly the same length, which is determined by the sub-trajectory with the maximum length in the batch. Notice here that if the batch size is sufficiently small, then all the sequences in the same batch have similar lengths, thus padding is not excessive and a lot of useless computation is avoided. 

\noindent\textbf{4) Batch feeding process:} Finally, the batches are fed to the network in a random order, so that the network doesn't overfit to certain trajectory lengths. This is a way to artificially enforce the learned representations to be length invariant. 

We observed that due to the relatively large length of many input trajectories (usually a few hundred time-steps), the Decoder was facing difficulties in generating accurate copies of the input by using only the context vector. Thus, given the fact that the reconstruction error was really important to be small throughout the entire normal dataset, we exploited \textbf{teacher forcing} \cite{Goodfellow-et-al-2016} to sufficiently reduce the reconstruction error. In our case, we don't use the predicted value as input to the Decoder, but the actual value of the time series at this specific time step (teacher forcing), to prevent error accumulation. We also reverse the desired output sequence, similar to \cite{Implement2}, to reduce the necessity of modeling very long-term dependencies between the initial and final values of the time series. 

In terms of optimization, the batch size used is 64, the number of epochs is 200, and the learning rate is initialized to 0.01 and reduced after four consecutive failures to decrease validation loss by a factor of 0.9. Finally, the optimizer chosen was RMSprop.  Training the system needs about 2.5 hours to train on a NVIDIA GeForce GTX1060 6GB GPU.

\subsection{Detection Stage}
The input of this stage are new, unseen trajectories. These trajectories are processed real-time with an on-line strategy. As a pre-processing step, the trajectories are split to sub-trajectories, similar to the trajectories used for training. Then, they are fed to the autoencoder in order to obtain their reconstructions errors $\epsilon$. An alert, indicating anomaly detection, is generated if $\epsilon$ exceeds a threshold denoted here as $\theta$, \ie $\epsilon > \theta$. A real-time execution of the system on data provided by the ISL iCrowd simulator can be found in the Supplementary Material.

For the determination of $\theta$, a validation set equipped with ground truth labels is used in order to calculate the reconstruction errors of both normal and abnormal patterns. We treat the problem as binary classification (normal vs abnormal) and determine $\theta$ by maximizing the $F_1$ measure of the abnormal class.

\section{Evaluation}
\label{Sec: Results}

\subsection{Dataset}
A synthetic dataset created using the ISL iCrowd simulator \cite{Scatc,iCrowd}, was used for evaluation. The dataset will be made publicly available for research purposes. 

\begin{figure} [htb]
  \centering
  \includegraphics[scale=0.32]{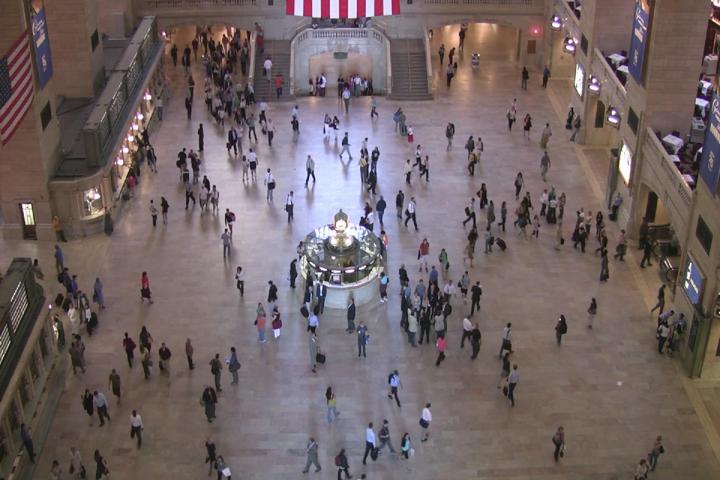}
  \caption{Real-world security footage.}
  \label{Fig: RealData}
\end{figure}

\begin{figure} [htb]
  \centering
  \includegraphics[scale=0.12]{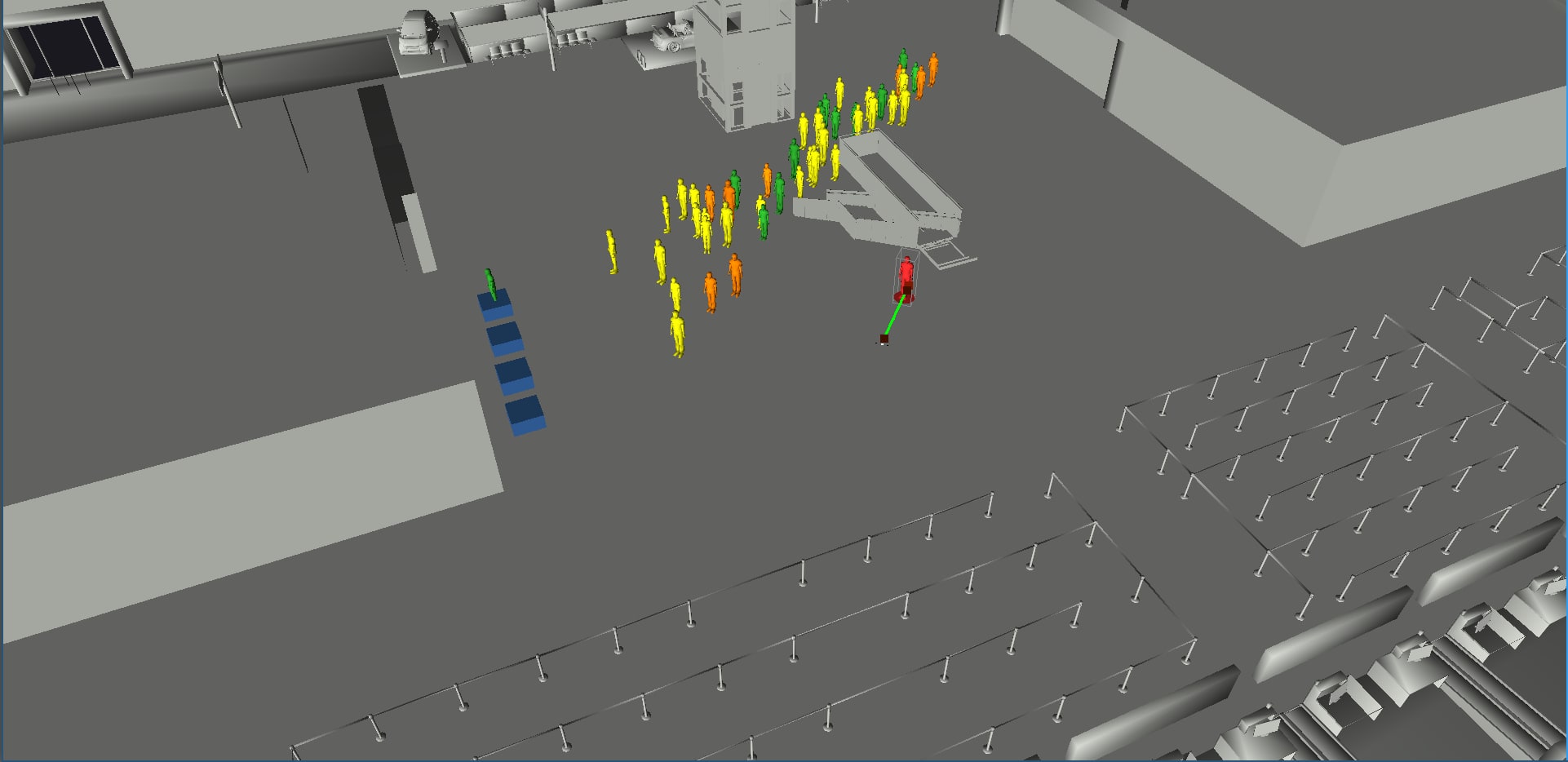}
  \caption{iCrowd-produced security footage illustrating normal trajectories and a specific abnormal trajectory pattern.}
  \label{Fig: SimDataAnom}
\end{figure}

\begin{figure} [h]
  \centering
  \includegraphics[width=\linewidth]{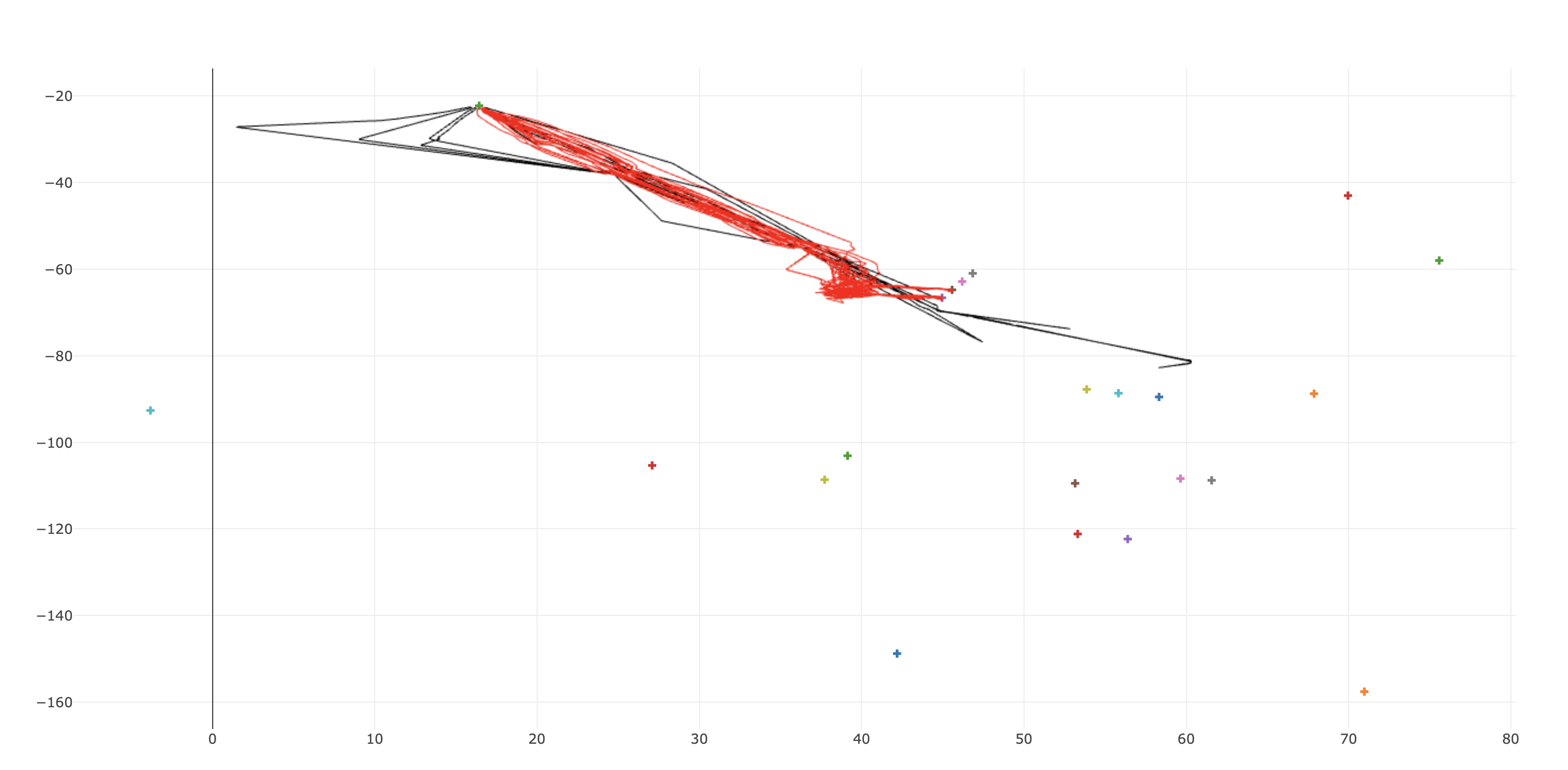}
  \caption{Exemplary normal (red) and abnormal (black) trajectories from the iCrowd dataset. Please see color version for better visibility.}
  \label{Fig: Dataset1Ex}
\end{figure}

\textbf{ISL iCrowd Dataset:} 
ISL iCrowd can be utilized as a simulator generating the positions of the living entities using its simulation engine \cite{Scatc,iCrowd}, as well as a visualizer displaying their positions. Figure \ref{Fig: RealData} illustrates a real-world security footage which can be directly compared to Figure \ref{Fig: SimDataAnom} which illustrate iCrowd-produced security footage. The similarity regarding the position of the pedestrians is obvious. An exemplary subset of the normal and abnormal trajectories is shown in Figure \ref{Fig: Dataset1Ex}. As can be seen, sometimes abnormal patterns can be very similar to normal ones, making the task quite challenging.

The dataset used to train the network was produced by simulating a realistic scenario of crowd behavior in a small airport. In particular, 20 flight departures were simulated, randomly distributed in a specific time interval in order to increase congestion variations in various parts of the airport, thus making the conditions of the simulation closer to reality. Each flight carried 50 different passengers that enter the airport at random times before their flight departs. Please see
supplementary material for videos of the simulation. 

\textbf{Benchmarks:} 
In order to quantitatively assess our system we run the simulation under 3 different scenarios to generate normal trajectories: a) Small Congestion: 400 passengers distributed in 20 flights, b) Normal Congestion: 1000 passengers distributed in 20 flights; c) Large Congestion: 4000 passengers distributed in 20 flights. All the entities are behaving in a normal way throughout their entire trajectories inside the airport.

\begin{figure*}[t]
\centering
    \begin{subfigure}{0.5\linewidth}
    \includegraphics[scale=0.4]{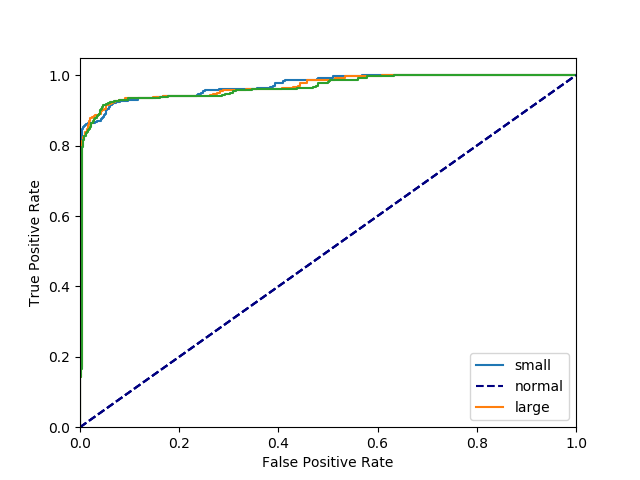}
    \label{fig:roc}
    \end{subfigure}\hfill
    \begin{subfigure}{0.5\linewidth}
    \includegraphics[scale = 0.4]{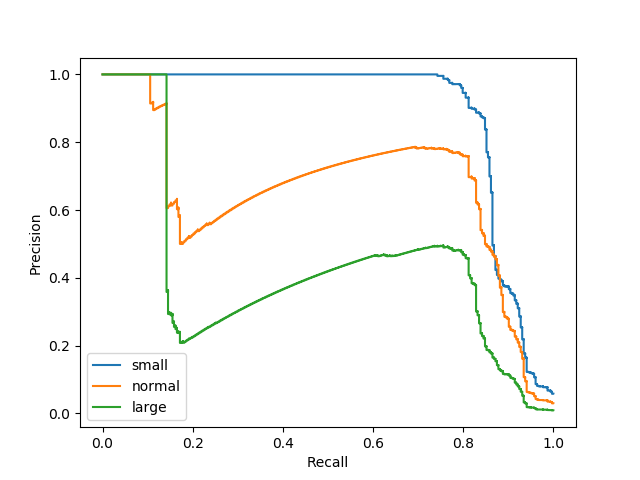}
    \label{fig:pr}
    \end{subfigure}
    \caption{Left: ROC curves of the 3 benchmarks, Right: PR curves.}    \label{fig:curves}
\end{figure*}

We also artificially create a dataset with abnormal trajectories by simulating 300 passengers distributed in 20 flights that perform various different abnormal movements in specific parts of the airport. We concatenate the abnormal dataset with each one of the normals to create our 3 different benchmarks and test the generalization abilities of our system to arbitrary conditions.

Note here, that the simulation may allow for video-based anomaly detection by deploying a series of systems in order to extract the trajectories (multi-person tracker, data association algorithms, denoising etc.). For this work, we assume that all the people are already detected and tracked, in order to assess the efficiency of our system alone. Thus, we directly use the trajectories given in the ISL iCrowd dataset. 

\subsection{Results}
\subsubsection{Quantitative Analysis}
For each one of the benchmarks we provide the \textbf{Precision-Recall (PR)} and \textbf{Receiver Operating Characteristics (ROC)} curves in Figure \ref{fig:curves}. Table \ref{table_metrics} provides the Average Precision (AP) and Area Under Curve of the ROC curve (AUC-ROC) scores respectively.

\begin{table}[t]
\begin{center}
\scalebox{0.85}{
\begin{tabular}{|l|c|c|c|}
\hline
\textbf{Benchmark} & \textbf{small}&\textbf{normal}&\textbf{large}\\\hline
\textbf{AP}&0.89&0.66&0.43\\\hline
\textbf{AUC-ROC}&0.97&0.97&0.97 \\ \hline
\textbf{F1 for $\theta = 4$}&0.87&0.78&0.60 \\ \hline
\textbf{accuracy for $\theta = 4$}&0.99&0.99&0.99 \\ \hline
\end{tabular}}
\end{center}
\caption{Metrics across the 3 benchmarks.}
\label{table_metrics}
\end{table}

\begin{figure}[h]
    \includegraphics[width=\linewidth]
     {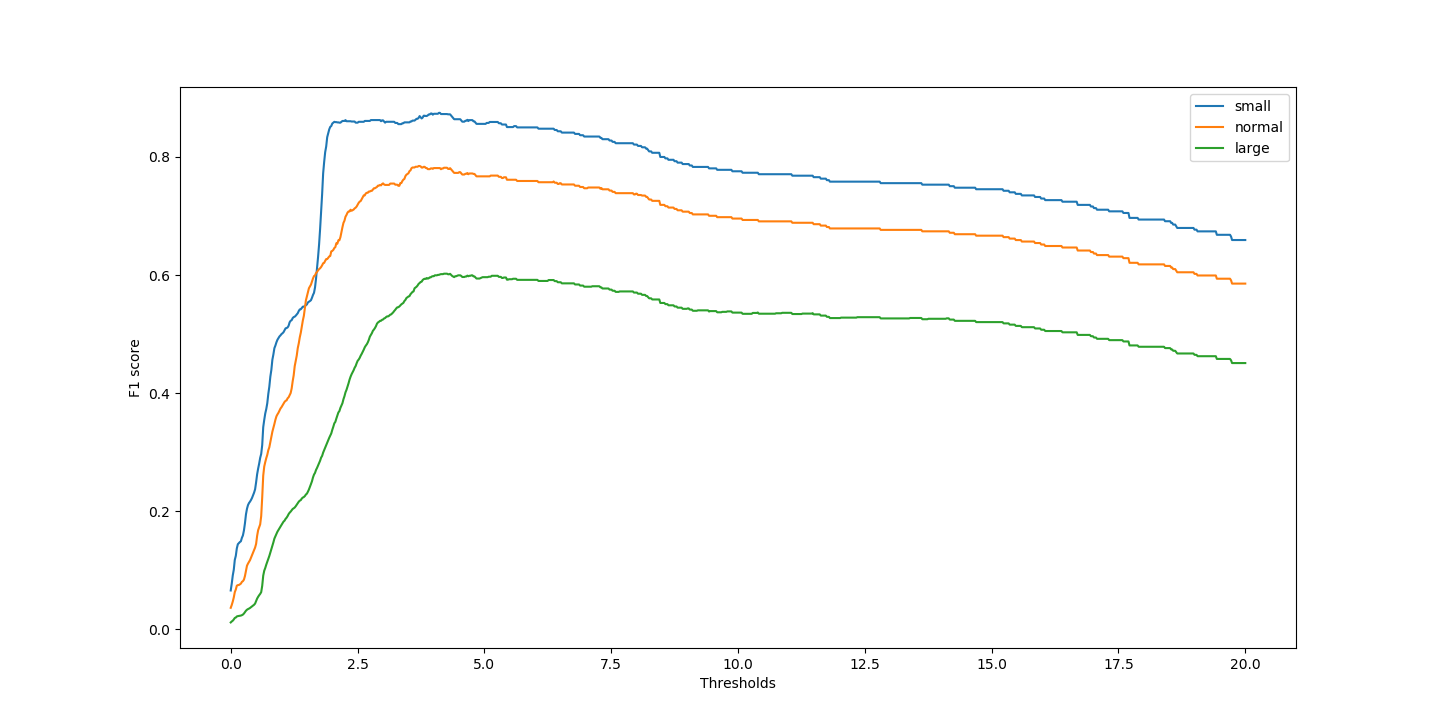}
    \caption{F1 vs thresholds curve.}
    \label{fig:f1}
\end{figure}

We also provide an analysis of the effect that the threshold has in the F1 score of the abnormal class in Figure \ref{fig:f1}. We use the F1 metric because the classes are highly imbalanced, making metrics such as accuracy less informative. As can be seen from the graph, our approach is highly robust w.r.t the threshold, regardless of the dataset, namely for thresholds close to 4 and above, the F1 score decreases at a quite slow rate. Note also that the thresholds that achieve the best results are quite similar across different datasets, which indicates that our system is less prone to hyperparameter tuning. In each of the datasets we found the threshold that maximizes the F1 score (4.1, 3.7, 4.2 for the small, normal and large congestion datasets, respectively). 

Finally, we chose a specific threshold and measured the performance of our system in terms of F1, precision and recall for both classes. Given the above, a reasonable choice for $\theta$ is 4. The results are provided in Tables \ref{table_small}, \ref{table_normal} and \ref{table_large}. 
\begin{table}[h]
\begin{center}
\scalebox{0.85}{
\begin{tabular}{|l|c|c|c|c|}
\hline
\textbf{Benchmark} & \textbf{precision}&\textbf{recall}&\textbf{F1-score}&\textbf{support}\\\hline
\textbf{normal} &0.993     &0.999     &0.996      &8646\\ \hline
\textbf{abnormal}    &0.968     &0.795     &0.873       &303\\ \hline
\textbf{average score}  &0.992     &0.992     &0.992      &8949 \\ \hline
\end{tabular}}
\end{center}
\caption{Metrics table for the small congestion benchmark.}
\label{table_small}
\end{table}

\begin{table}[h]
\begin{center}
\scalebox{0.85}{
\begin{tabular}{|l|c|c|c|c|}
\hline
\textbf{Benchmark} & \textbf{precision}&\textbf{recall}&\textbf{F1-score}&\textbf{support}\\\hline
\textbf{normal}  &0.996     &0.995     &0.996     &16140\\ \hline
\textbf{abnormal}    &0.768     &0.795     &0.781       &303\\ \hline
\textbf{average score}  &0.992     &0.992     &0.992     &16443 \\ \hline
\end{tabular}}
\end{center}
\caption{Metrics table for the normal congestion benchmark.}
\label{table_normal}
\end{table}

\begin{table}[h]
\begin{center}
\scalebox{0.85}{
\begin{tabular}{|l|c|c|c|c|}
\hline
\textbf{Benchmark} & \textbf{precision}&\textbf{recall}&\textbf{F1-score}&\textbf{support}\\\hline
\textbf{normal}  &0.999     &0.995     &0.997     &52612\\ \hline
\textbf{abnormal}    &0.479     &0.795     &0.598       &303\\ \hline
\textbf{average score}  &0.996     &0.994     &0.995     &52915 \\ \hline
\end{tabular}}
\end{center}
\caption{Metrics table for the large congestion benchmark.}
\label{table_large}
\end{table}

\subsubsection{Qualitative Analysis}
\begin{figure}[h]
    \centering
    \includegraphics[width=\linewidth]{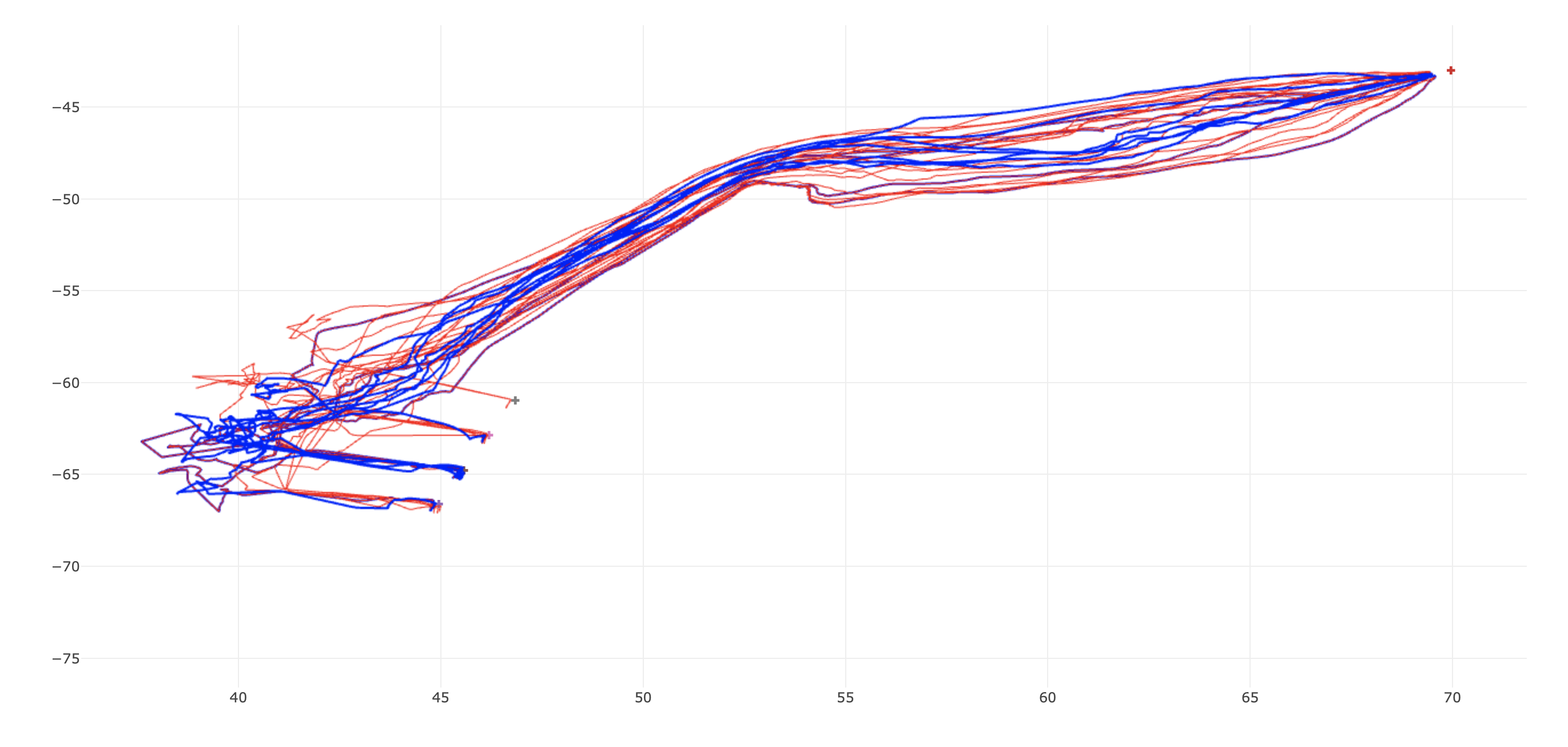}
     \caption{Visual comparison of correctly classified normal (red) trajectories to other normal (blue) that have been misclassifed as anomalies. Please see color version for better visibility.}
     \label{fig:false_positives}
\end{figure}

Notice that as the congestion increases the trajectories generated by the simulator can become highly unpredictable, \ie many passengers choose to move in ways that do not resemble the normal trajectories that the model has learned, thus leading the autoencoder to misclassify some normal trajectories as anomalies (see Figure \ref{fig:false_positives}). This is an expected behavior since they are potentially drawn from a part of the trajectory distribution that the autoencoder has not seen during train. This motivates us to create a model (as future work) that will be handling the anomalies as contextual ones, namely in order to classify a trajectory, one need to take into account contextual information (e.g. trajectories of neighboring passengers).  Nevertheless, the number of false positives is quite low compared to the total number of trajectories that the system has to classify. 

\begin{figure}[h]
    \centering
    \includegraphics[width=\linewidth]{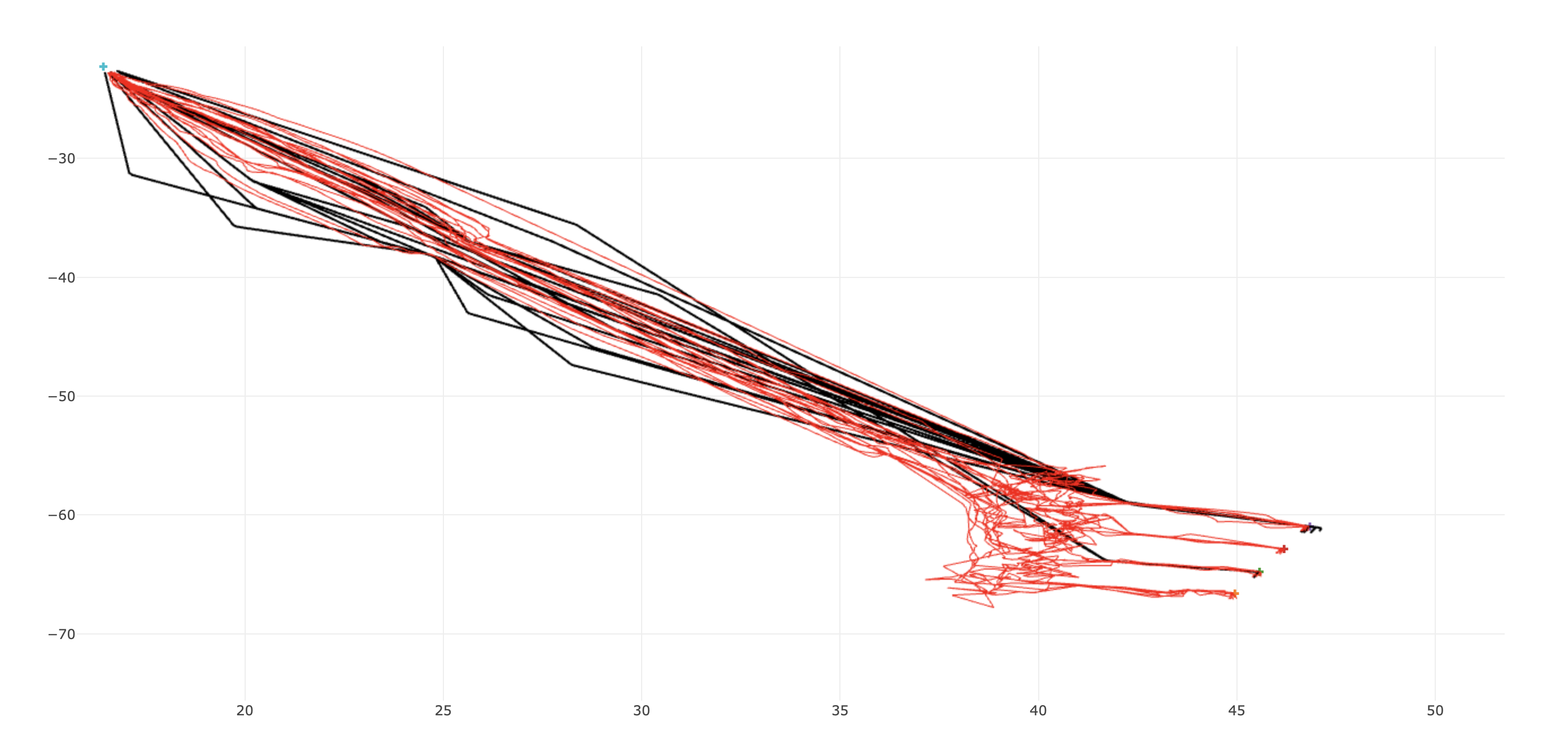}
     \caption{Visual comparison of certain normal (red) trajectories to some anomalies (black) that have been misclassifed as normal. Please see color version for better visibility.}
     \label{fig:false_negatives}
\end{figure}

Regarding the anomalies that the system failed to predict, we performed a qualitative analysis by plotting a random sample and inspecting the shapes of the trajectories (Figure \ref{fig:false_negatives}). It is clear that these trajectories are difficult to be distinguished even by a human assessor. Soft thresholding could be used, namely providing an anomaly score rather than an anomaly label, in order to raise alerts (the risk warning will be scaled according to the anomaly score) for a human annotator/supervisor. 
This can also be extended through an active learning-scheme, where the human annotator will be labeling examples that the classifier cannot predict with strong confidence. These examples will then be reused by the classifier, in order to improve its performance (mainly regarding cases near the decision boundary).

\begin{figure}[h]
    \centering
    \includegraphics[width=\linewidth]{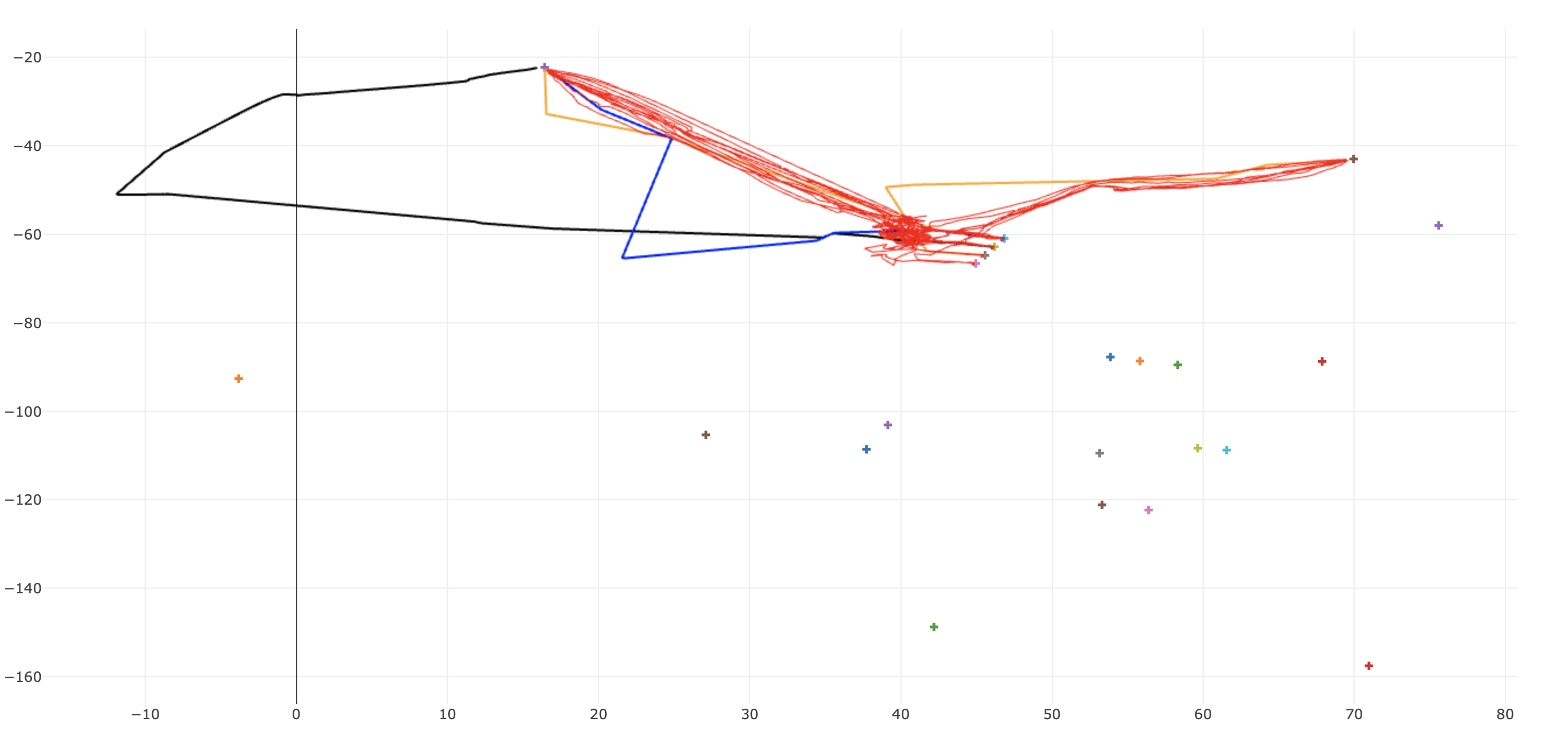}
     \caption{Visual comparison of correctly classified normal (red) trajectories to correctly classified anomalies. Orange color shows errors between 4 and 10, blue between 10 and 30 and black larger than 30. Please see color version for better visibility.}
     \label{fig:true_positives}
\end{figure}

Figure \ref{fig:true_positives} illustrates some of the true positives of the dataset. As can be clearly seen, the values of the errors agree with the human intuition about the level of suspiciousness of each trajectory. Thus, our predictions can go beyond binary classification, and additionally work as a measure of the probability of a trajectory being anomalous.

\begin{figure}[h]
    \centering
    \includegraphics[scale=0.2]{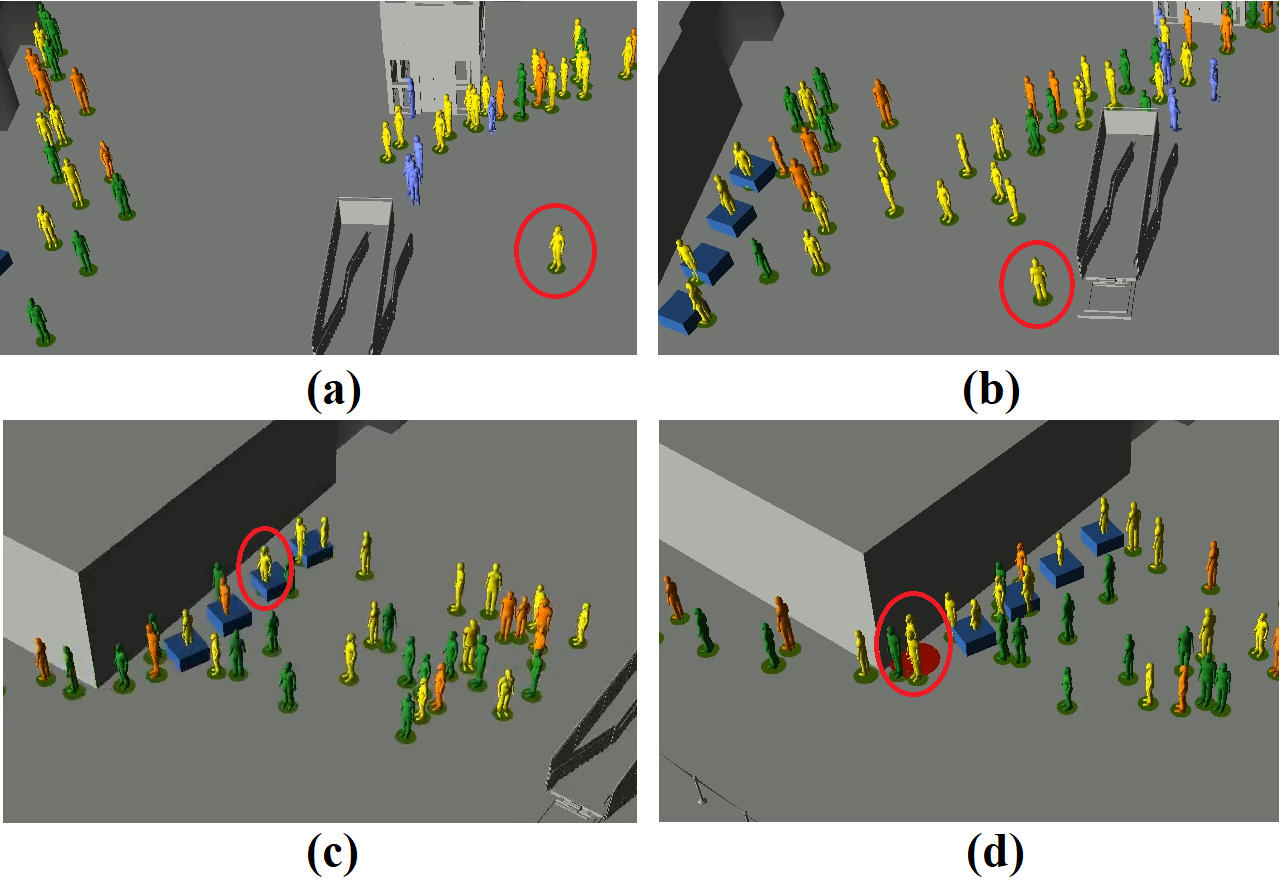}
     \caption{Real-time predictions by the proposed method.}
     \label{Fig: VideoScreenshots}
\end{figure}

Finally, Figure \ref{Fig: VideoScreenshots} illustrates consecutive frames of a real-time exhibition video of the proposed method. The pedestrian in the red circle follows a declined trajectory in sub-figures (a) and (b). In sub-figure (c) they pass through the control point and a few seconds later, in sub-figure (d) they are characterized as suspicious. Every passenger has an individual circle underneath which indicates the corresponding anomaly score. Green implies a normal trajectory, while red a suspiciously declined one. More exhibition videos are available in the supplementary material.

\section{Future Challenges}
\label{Sec: FutureChallenges}
In the future, the current research is planned to be extended towards several directions. Our scheme is planned to be tested in publicly available datasets, such as the one presented in \cite{Piciarelli} as well as in datasets containing noisy data, where the anomalies are more subtle and sophisticated. 

In its current form, the presented work uses the position coordinates of the trajectories as input features. In the future, additional features such as the velocity, acceleration and heading of the pedestrian as well as the Fourier coefficients of the initial trajectory time series for each coordinate could be extracted and recruited. A dedicated LSTM auto-encoder will be utilised for training on each feature and fusion methodologies, such as the one in \cite{Fusion}, will be used for extracting a single output out of the the entire cluster of LSTM networks.

Finally, alternative architectures could be tested, such as GANs \cite{NIPS_GANS}, and hierarchical Sequence to Sequence models (which could tackle the problem of local and global anomalies jointly) and architectures that account for contextual anomalies \cite{SocialDL}.

\section{Conclusion}
\label{Sec: Conclusions}
In this paper, a fully automated technique for real-time detection of anomalies in human trajectories is proposed. The algorithm is based on a Sequence to Sequence LSTM auto-encoder. The detection scheme was tested on a synthetic dataset created by the iCrowd human simulator. With this work we provide insights regarding the application of deep learning in modeling human motion patterns captured by surveillance sensors. Our results imply that deviations from normal behaviors can be accurately predicted, thus being an important proof of concept in terms of the use of deep learning in safety critical applications. 

\section{Acknowledgments}
The research described in this paper has been supported by the following research contracts: “FLYSEC: Optimizing time-to-FLY and enhancing airport SECurity,” Programme: Horizon 2020, European Union Grant Agreement No. 653879, Duration: 01/05/2015 - 31/07/2018, \url{http://www.fly-sec.eu}, and “TRESSPASS: Robust Risk Based Screening and Alert System for Passengers and luggage,” Grant Agreement No. 787120, Call: H2020-SEC-2016-2017-2, \url{https://www.tresspass.eu/The-project}.

{\small
\bibliographystyle{ieee}
\bibliography{AVSS_2019.bib}

\begin{thebibliography}{10}\itemsep=-1pt

\bibitem{SocialDL}
A.~Alahi, K.~Goel, V.~Ramanathan, A.~Robicquet, L.~Fei-Fei, and S.~Savarese.
\newblock {Social LSTM: Human Trajectory Prediction in Crowded Spaces}.
\newblock In {\em 2016 IEEE Conference on Computer Vision and Pattern
  Recognition (CVPR)}, pages 961--971, June 2016.

\bibitem{Bu_etal_anomaly_monitoring_trajectory_streams_KDD2009}
Y.~Bu, L.~Chen, A.~W.-C. Fu, and D.~Liu.
\newblock Efficient anomaly monitoring over moving object trajectory streams.
\newblock In {\em Proceedings of the 15th ACM SIGKDD international conference
  on Knowledge discovery and data mining}, pages 159--168. ACM, 2009.

\bibitem{Chen_etal_iBOAT_TITN2013}
C.~Chen, D.~Zhang, P.~S. Castro, N.~Li, L.~Sun, S.~Li, and Z.~Wang.
\newblock iboat: Isolation-based online anomalous trajectory detection.
\newblock {\em IEEE Transactions on Intelligent Transportation Systems},
  14(2):806--818, 2013.

\bibitem{Implement1}
K.~Cho, B.~van Merri{\"{e}}nboer, {\c C}.~G{\"{u}}l{\c c}ehre, D.~Bahdanau,
  F.~Bougares, H.~Schwenk, and Y.~Bengio.
\newblock {Learning Phrase Representations using RNN Encoder--Decoder for
  Statistical Machine Translation}.
\newblock In {\em Proceedings of the 2014 Conference on Empirical Methods in
  Natural Language Processing (EMNLP)}, pages 1724--1734. Association for
  Computational Linguistics, 2014.

\bibitem{Fu_SIMILARITY_VEHICLE_TRAJECTORY_CLUSTERING_ANOMALY_ICIP2005}
Z.~Fu, W.~Hu, and T.~Tan.
\newblock Similarity based vehicle trajectory clustering and anomaly detection.
\newblock In {\em Image Processing, 2005. ICIP 2005. IEEE International
  Conference on}, volume~2, pages II--602. IEEE, 2005.

\bibitem{Ge_et_al_TOP-EYE_CIKM2010}
Y.~Ge, H.~Xiong, Z.-h. Zhou, H.~Ozdemir, J.~Yu, and K.~C. Lee.
\newblock Top-eye: Top-k evolving trajectory outlier detection.
\newblock In {\em Proceedings of the 19th ACM international conference on
  Information and knowledge management}, pages 1733--1736. ACM, 2010.

\bibitem{Goodfellow-et-al-2016}
I.~Goodfellow, Y.~Bengio, and A.~Courville.
\newblock {\em Deep Learning}.
\newblock MIT Press, 2016.
\newblock \url{http://www.deeplearningbook.org}.

\bibitem{NIPS_GANS}
I.~Goodfellow, J.~Pouget-Abadie, M.~Mirza, B.~Xu, D.~Warde-Farley, S.~Ozair,
  A.~Courville, and Y.~Bengio.
\newblock Generative adversarial nets.
\newblock In {\em Advances in Neural Information Processing Systems 27}, pages
  2672--2680. 2014.

\bibitem{hochreiter1997long}
S.~Hochreiter and J.~Schmidhuber.
\newblock Long short-term memory.
\newblock {\em Neural computation}, 9(8):1735--1780, 1997.

\bibitem{Johansson_Falkman_vessel_anomalies_Bayesian_network_ISSNIP_2007}
F.~Johansson and G.~Falkman.
\newblock Detection of vessel anomalies-a bayesian network approach.
\newblock In {\em Intelligent Sensors, Sensor Networks and Information, 2007.
  ISSNIP 2007. 3rd International Conference on}, pages 395--400. IEEE, 2007.

\bibitem{Scatc}
V.~Kountouriotis, S.~C.~A. {Thomopoulos}, and Y.~Papelis.
\newblock An agent-based crowd behaviour model for real time crowd behaviour
  simulation.
\newblock {\em Pattern Recognition Letters}, 44:30 -- 38, 2014.
\newblock Pattern Recognition and Crowd Analysis.

\bibitem{iCrowd}
V.~I. Kountouriotis, M.~Paterakis, and S.~C.~A. {Thomopoulos}.
\newblock {iCrowd: agent-based behavior modeling and crowd simulator}.
\newblock In {\em Signal Processing, Sensor/Information Fusion, and Target
  Recognition XXV}, volume 9842, page 98420Q. International Society for Optics
  and Photonics, 2016.

\bibitem{Laxhammar_Falkman_Conformal_Distribution-Independent_Anomaly_Streaming_StreamKDD2010}
R.~Laxhammar and G.~Falkman.
\newblock Conformal prediction for distribution-independent anomaly detection
  in streaming vessel data.
\newblock In {\em Proceedings of the First International Workshop on Novel Data
  Stream Pattern Mining Techniques}, StreamKDD '10, pages 47--55, New York, NY,
  USA, 2010. ACM.

\bibitem{Laxhammar_Falkman_Conformal_Anomaly_Haussdorf_ICIF2011}
R.~Laxhammar and G.~Falkman.
\newblock Sequential conformal anomaly detection in trajectories based on
  hausdorff distance.
\newblock In {\em Information Fusion (FUSION), 2011 Proceedings of the 14th
  International Conference on}, pages 1--8. IEEE, 2011.

\bibitem{Laxhammar1}
R.~Laxhammar and G.~Falkman.
\newblock Online learning and sequential anomaly detection in trajectories.
\newblock {\em IEEE Transactions on Pattern Analysis and Machine Intelligence},
  36(6):1158--1173, June 2014.

\bibitem{Laxhammar_etal_anomaly_sea_traffic_comparison_GMM_KDE_ICIF_2009}
R.~Laxhammar, G.~Falkman, and E.~Sviestins.
\newblock Anomaly detection in sea traffic-a comparison of the gaussian mixture
  model and the kernel density estimator.
\newblock In {\em Information Fusion, 2009. FUSION'09. 12th International
  Conference on}, pages 756--763. IEEE, 2009.

\bibitem{Lee_trajectory_outlier_partition_detect_ICDE2008}
J.~G. Lee, J.~Han, and X.~Li.
\newblock Trajectory outlier detection: A partition-and-detect framework.
\newblock In {\em 2008 IEEE 24th International Conference on Data Engineering},
  pages 140--149, April 2008.

\bibitem{Li_et_al_ROAM_SDM_2007}
X.~Li, J.~Han, S.~Kim, and H.~Gonzalez.
\newblock Roam: Rule- and motif-based anomaly detection in massive moving
  object data sets.
\newblock In {\em In Proceedings of 7th SIAM International Conference on Data
  Mining}, 2007.

\bibitem{Li_etal_Temporal_outlier_vehicle_traffic_ICDE_2009}
X.~Li, Z.~Li, J.~Han, and J.~G. Lee.
\newblock Temporal outlier detection in vehicle traffic data.
\newblock In {\em 2009 IEEE 25th International Conference on Data Engineering},
  pages 1319--1322, March 2009.

\bibitem{Liao_etal_anomaly_GPS_visual_analytics_VAST2010}
Z.~Liao, Y.~Yu, and B.~Chen.
\newblock Anomaly detection in gps data based on visual analytics.
\newblock In {\em 2010 IEEE Symposium on Visual Analytics Science and
  Technology}, pages 51--58, Oct 2010.

\bibitem{newwork2}
C.~Ma, Z.~Miao, M.~Li, S.~Song, and M.-H. Yang.
\newblock Detecting anomalous trajectories via recurrent neural networks.
\newblock 2018.

\bibitem{Piciarelli}
C.~Piciarelli, C.~Micheloni, and G.~L. Foresti.
\newblock {Trajectory-Based Anomalous Event Detection}.
\newblock {\em IEEE Transactions on Circuits and Systems for Video Technology},
  18(11):1544--1554, Nov 2008.

\bibitem{Fusion}
C.~Rizogiannis, K.~G. Thanos, A.~Astyakopoulos, D.~M. Kyriazanos, and S.~C.~A.
  Thomopoulos.
\newblock Sensor data monitoring and decision level fusion scheme for early
  fire detection.
\newblock In {\em SPIE Signal Processing, Sensor/Information Fusion, and Target
  Recognition XXVI}, volume 10200, 2017.

\bibitem{Silito_FIsher_Semi-supervised_Anomalous_Trajectory_Detection_BMCV2008}
R.~R. Sillito and R.~B. Fisher.
\newblock Semi-supervised learning for anomalous trajectory detection.
\newblock In {\em In Proc. BMVC}, pages 1035--1044, 2008.

\bibitem{Matr1}
I.~Sutskever, O.~Vinyals, and Q.~V. Le.
\newblock Sequence to sequence learning with neural networks.
\newblock In {\em Advances in neural information processing systems}, pages
  3104--3112, 2014.

\bibitem{Implement2}
I.~Sutskever, O.~Vinyals, and Q.~V. Le.
\newblock {Sequence to Sequence Learning with Neural Networks}.
\newblock In {\em Advances in Neural Information Processing Systems 27}, pages
  3104--3112. Curran Associates, Inc., 2014.

\bibitem{Suzuki_etal_motion_patterns_anomaly_human_trajectory_ICSMC2007pdf}
N.~Suzuki, K.~Hirasawa, K.~Tanaka, Y.~Kobayashi, Y.~Sato, and Y.~Fujino.
\newblock Learning motion patterns and anomaly detection by human trajectory
  analysis.
\newblock In {\em 2007 IEEE International Conference on Systems, Man and
  Cybernetics}, pages 498--503, Oct 2007.

\bibitem{FINAL}
S.~C.~A. Thomopoulos, S.~Daveas, and A.~Danelakis.
\newblock Automated real-time risk assessment for airport passengers using a
  deep learning architecture.
\newblock In {\em SPIE Defense + Commercial Sensing, Signal Processing,
  Sensor/Information Fusion, and Target Recognition XXVIII}, volume 11018,
  2019.

\bibitem{Wang_et_al_Learning_Semantic_Scene_Models_ECCV2006}
X.~Wang, K.~Tieu, and E.~Grimson.
\newblock Learning semantic scene models by trajectory analysis.
\newblock In {\em Proceedings of the 9th European Conference on Computer Vision
  - Volume Part III}, ECCV'06, pages 110--123, Berlin, Heidelberg, 2006.
  Springer-Verlag.

\bibitem{Matr2}
Y.~Wu, M.~Schuster, Z.~Chen, Q.~V. Le, M.~Norouzi, W.~Macherey, M.~Krikun,
  Y.~Cao, Q.~Gao, K.~Macherey, et~al.
\newblock Google's neural machine translation system: Bridging the gap between
  human and machine translation.
\newblock {\em arXiv preprint arXiv:1609.08144}, 2016.

\bibitem{newwork1}
D.~Yao, C.~Zhang, Z.~Zhu, J.~Huang, and J.~Bi.
\newblock Trajectory clustering via deep representation learning.
\newblock In {\em 2017 International Joint Conference on Neural Networks
  (IJCNN)}, pages 3880--3887, May 2017.

\bibitem{Zhang_etal_iBAT_UbiComp2011}
D.~Zhang, N.~Li, Z.-H. Zhou, C.~Chen, L.~Sun, and S.~Li.
\newblock ibat: Detecting anomalous taxi trajectories from gps traces.
\newblock In {\em Proceedings of the 13th International Conference on
  Ubiquitous Computing}, UbiComp '11, pages 99--108, New York, NY, USA, 2011.
  ACM.

\end{thebibliography}
}

\end{document}